\documentclass{article}
\usepackage{spconf,amsmath,graphicx,hyperref,booktabs,amssymb}
\usepackage{multirow}
\usepackage{xcolor}
\title{Adaptive Overclocking: Dynamic Control of Thinking Path Length via Real-Time Reasoning Signals}
\name{\begin{tabular}[t]{c@{}c@{}}
Shuhao Jiang$^{*\dagger}$\thanks{$^*$These authors contributed equally to this work and should be considered co-first authors.}\thanks{$^\dagger$Corresponding author.},
      Songbo Wang$^*$, 
      Yang Qiao, 
      Chun Xu, 
      Chaoyang Zheng, 
      Shengyi Zhou, & \\
      Huanjun Wang, 
      Fangming Li, 
      Cong Zhang, 
      Jiyu Wang
      \end{tabular}
      }
\address{Huawei Technologies Co., Ltd., Dongguan, China}

\begin{document}
\ninept
\maketitle

\begin{abstract}
Large Reasoning Models (LRMs) often suffer from computational inefficiency due to \textit{overthinking}, where a fixed reasoning budget fails to match the varying complexity of tasks. To address this issue, we propose \textit{Adaptive Overclocking}, a method that makes the overclocking hyperparameter $\alpha$ dynamic and context-aware. Our method adjusts reasoning speed in real time through two complementary signals: (1) token-level model uncertainty for fine-grained step-wise control, and (2) input complexity estimation for informed initialization. We implement this approach with three strategies: Uncertainty-Aware Alpha Scheduling (UA-$\alpha$S), Complexity-Guided Alpha Initialization (CG-$\alpha$I), and a Hybrid Adaptive Control (HAC) that combines both. Experiments on GSM8K, MATH, and SVAMP show that HAC achieves superior accuracy–latency trade-offs, reducing unnecessary computation on simple problems while allocating more resources to challenging ones. By mitigating overthinking, Adaptive Overclocking enhances both efficiency and overall reasoning performance.

\end{abstract}

\section{Introduction}
\label{sec:intro}

Large Language Models (LLMs) such as GPT-O1\cite{o1} and DeepSeek-R1\cite{guo2025deepseek} have demonstrated remarkable performance in complex reasoning tasks methods such as Chain-of-Thought (CoT) prompting\cite{wei2023chainofthoughtpromptingelicitsreasoning}, which encourages models to generate explicit intermediate reasoning steps before reaching  final answer. This structured approach has proven to be fundamental to improving model performance in domains like mathematics, logic, and code generation. However, this advancement presents a critical trade-off. While more elaborate reasoning chains can improve accuracy on complex problems, they often introduce significant computational overhead, increased inference latency, and a phenomenon known as ``overthinking''.\cite{cuadron2025danger}
Overthinking occurs when a model generates verbose, redundant, or even counterproductive reasoning steps, which makes LLMs suffers from waste computational resources and degrade the quality of the final answer as well.\cite{cuadron2025danger} This establishes a pressing need for methods that can promote reasoning efficiency, characterized by optimizing the length and quality of the thinking process to achieve a balance between accuracy and efficiency.\cite{wu2025more}

There has been a lot of interests in solving overthinking problems. Eisenstadt et al. introduced
a paradigm shift in controlling model reasoning with their ``Overclocking'' method.\cite{Eisenstadt2025Overclocking} Their work diverged from conventional approaches that focus on prompt engineering or output post-processing. Instead, they demonstrated that it is possible to directly manipulate a model's internal hidden states to control the length of its reasoning path. They identified a directional vector within the model's activation space, termed the Thinking Progress Vector (TPV), which encodes the model's internal estimate of its progress through a reasoning task. By applying a  static, constant intervention along this vector during inference, they could effectively accelerate the model's thinking process, compelling it to reach a conclusion more quickly.

Despite its novelty, the static nature of this intervention presents a significant limitation. Reasoning is not a monolithic process, but consists of steps with varying complexity, ranging from routine calculations to complex logical leaps. A static intervention is inherently inflexible, applying the same degree of acceleration indiscriminately. This approach can be suboptimal, potentially rushing the model through critical steps where careful deliberation is needed, or failing to provide a sufficient push when the model is stuck in a repetitive loop. Such limitation points to a clear research gap: the need for a more intelligent adaptive control mechanism.

In this work, we introduce \textbf{Adaptive Overclocking}, a novel method that elevates the static intervention paradigm to a dynamic, closed-loop control system. Our approach replace the static intervention parameter with a dynamic function that modulates the acceleration strength at each generation step. This function is guided by real-time signals derived from the model’s reasoning state, applying the uncertainty of next-token prediction as an indicator of overclocking strength signals. To summarize, our primary contributions are as follows:
\begin{enumerate}
    \item We formalize an adaptive control mechanism for LLM reasoning that operates entirely at inference time without requiring model retraining, and can be seamlessly combined with many existing acceleration methods.  
    \item We identify and leverage real-time signals, together with Complexity-Guided Alpha Initialization (CG-$\alpha$I) and to enable dynamic, state-aware interventions directly within the model's generative process.  
    \item Extensive experiments show that our adaptive approach consistently outperforms baselines, including the original static overclocking method, in terms of both accuracy and efficiency.  
\end{enumerate}

\section{Related Work}

% {\bf The Reasoning Length Trade-off.} 
% Chain-of-Thought (CoT) prompting~\cite{wei2022chain} demonstrated that eliciting intermediate reasoning steps is crucial for solving complex tasks. There has been lots of recent works discussed the trade-off between the performance and the length of reasoning chain. On one hand, excessive or redundant reasoning steps can lead to "overthinking", where performance degrades due to error accumulation or deviation from the task, while also incurring unnecessary computational costs~\cite{su2025between, sui2503stop, chen2024not}. On the other hand, "underthinking", or premature termination of the reasoning process, prevents the model from capturing the necessary complexity to reach a correct solution~\cite{wang2025thoughts}. This dilemma highlights the urgent need for robust mechanisms to control the length and trajectory of LLM reasoning.

\textbf{The Reasoning Length Trade-off}. Chain-of-Thought (CoT) prompting~\cite{wei2022chain} reveals that intermediate reasoning is crucial for complex tasks. 
However, the length of the reasoning chain presents a critical trade-off. 
Excessive reasoning, or "overthinking," can degrade performance through error accumulation and increased computational costs~\cite{su2025between, sui2503stop, chen2024not}. 
Conversely, insufficient reasoning, or "underthinking," prevents the model from capturing the complexity required for a correct solution~\cite{wang2025thoughts}. 
This dilemma highlights the urgent need for mechanisms to control the length and trajectory of an LLM's reasoning.

Strategies for managing LLM reasoning length can be classified into training-time and inference-time interventions. 
While training-time methods modify model parameters directly, such as through reinforcement learning~\cite{aggarwal2025l1,arora2025traininglanguagemodelsreason}, our work focuses on inference-time approaches, which can be further distinguished as static or dynamic intervention methods. 
\textbf{Static interventions} apply a single and fixed rule throughout generation. A prime example is the \textbf{Overclocking} method~\cite{Eisenstadt2025Overclocking}, which uses a constant parameter to overclock the inference, thereby reducing the reasoning length but lacks the adaptability for problems of varying complexity. 
In contrast, \textbf{dynamic interventions} adjust their behavior in real-time. Existing dynamic methods, however, tend to be output-reactive. They intervene by observing the model's generated text and subsequently modifying prompts\cite{li2024escapeskyhighcostearlystopping} or injecting external guidance~\cite{kojima2022large, wu2025effectively, jin2024impact}. This approach offers more flexibility than static methods but does not leverage information about the model's internal computational process as it unfolds

\textbf{Metacognition and Mechanistic Interpretability}:
Metacognition, defined as the ability to monitor and regulate one's own thought processes, provides a conceptual blueprint for adaptive self-regulation~\cite{flavell1979metacognition, zimmerman2002becoming}. In cognitive science, it explains how humans dynamically allocate mental effort, accelerating when confident and proceeding cautiously when uncertain. This principle highlights the potential for AI systems to achieve more intelligent and resource-efficient reasoning.
Mechanistic Interpretability seeks to reverse-engineer neural networks to understand their internal computational mechanisms~\cite{elhage2021mathematical, olah2020zoom}. A core insight from this field is that a model's internal states contain meaningful, structured signals about its reasoning process. This provides the technical foundation to move beyond interventions based on final outputs to those that can actively control the model's internal operations in real-time.
Synthesizing these concepts, our work introduces Adaptive Overclocking, a novel control system that equips LLMs with a form of computational metacognition by leveraging internal state signals to dynamically regulate the reasoning process.
% Our method of using internal model signals is inspired by two key fields. The first is \textbf{metacognition}, which is the ability to self-monitor one's own thought processes~\cite{flavell1979metacognition, zimmerman2002becoming}. We give the LLM a computational version of this ability; by tracking signals like its own prediction confidence, the model can assess its internal state and adjust its reasoning effort accordingly---speeding up when confident and slowing down when uncertain.

% The second field is \textbf{mechanistic interpretability}, which aims to understand how neural networks work internally~\cite{elhage2021mathematical, olah2020zoom}. Research in this area has isolated circuits for specific behaviors like in-context learning~\cite{olsson2022context} and arithmetic~\cite{nanda2023progress}. While most of this work focuses on analysis, our work takes a practical, applied approach. We leverage the insights from interpretability---that internal states contain meaningful signals---not just to observe the model, but to actively control it. Our main contribution is using these signals to create a fine-grained, state-aware intervention system that operates during inference. This capability for real-time, internal control is what distinguishes our method from prior static~\cite{Eisenstadt2025Overclocking} and dynamic~\cite{wu2025effectively} control strategies.
\begin{figure*}[t]
  \centering
  \includegraphics[width=0.9\textwidth]{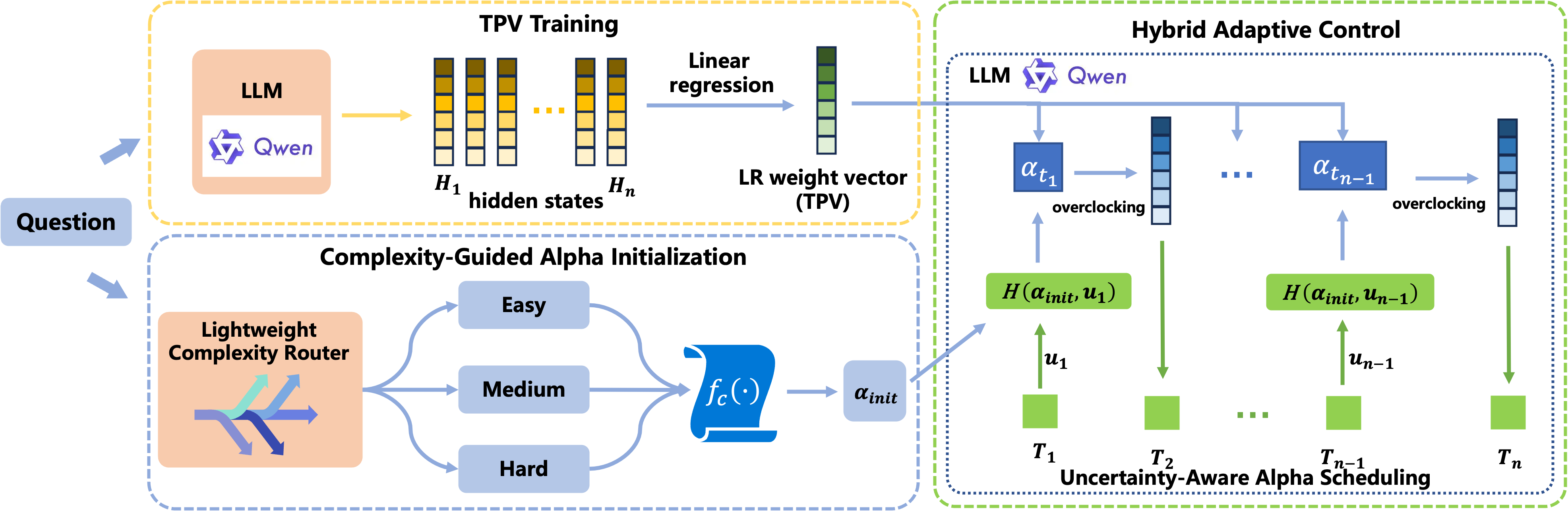}
    \caption{\textbf{Overall framework of Adaptive Overclocking.} 
  First, TPV is learned by regressing token positions from hidden states of reasoning trajectories, yielding a progress thinking direction. 
  During generation, hidden states are guided to a more direct path to the solution, where $\alpha_{t_{i}}$ controls the reasoning pace. 
  Build upon it, we further introduce two dynamic adaptive control strategies:
  (i) \textbf{ CG-$\alpha$I} assigns a global starting $\alpha_{init}$ based on input difficulty.
  and (ii)For every token $T_i$, \textbf{UA-$\alpha$S} further refines $\alpha$ token-by-token according to predictive uncertainty $u_i$, slowing down when entropy is high and speeding up when confidence is strong. 
  Finally, \textbf{HAC} unifies the two by using $\alpha_{\text{init}}$ from CG-$\alpha$I as the baseline and adaptively expanding it through UA-$\alpha$S, thus ensuring both global difficulty alignment and local uncertainty adaptation.}
  \label{fig:framework}
\end{figure*}

\section{Methods}
\label{sec:methods}
In this section, we present our adaptive overclocking framework, and Figure~\ref{fig:framework} provides a detailed illustration of our proposed architecture. Unlike previous static approaches, our method introduces a dynamic $\alpha$ control mechanism that transforms the system into a self-adaptive reasoning engine. We begin by reviewing the Thinking Progress Vector (TPV) intervention in \S~\ref{sec:tpv}, and then elaborate on our dynamic $\alpha$ control strategies in \S~\ref{sec:dynamic-alpha}, including Hybrid Adaptive Control , which unifies global initialization and local scheduling into a single mechanism.

\subsection{Thinking Progress Vector Intervention}
\label{sec:tpv}

We follow the intervention mechanism proposed by Eisenstadt et al.\cite{Eisenstadt2025Overclocking} as the backbone of our approach. The process begins by generating full reasoning trajectories with a base large reasoning model (e.g., DeepSeek-R1-Distill-Qwen32B\cite{guo2025deepseek}). Each trajectory contains the complete chain of thought from \texttt{<think>} to \texttt{</think>}.

For a trajectory of length $N$, let $h_j$ be the hidden state of the $j$-th token, paired with its normalized position $p_j = j/N$. This yields a dataset 
\begin{equation}
D = \{(h_j, p_j)\}_{j=1}^N ,
\end{equation}
on which we fit a linear regression model to predict $p_j$ from $h_j$. The regression vector $\theta$ is referred to as the \emph{Thinking Progress Vector} (TPV).  

A static intervention modifies the hidden state by
\begin{equation}\label{eq:static-intervention}
h_{\alpha} = h + \alpha \theta ,
\end{equation}
 where the parameter $\alpha$ scales the intervention. The magnitude of $\alpha$ determines the degree of intervention, with larger absolute values signifying a stronger modification. 
This intervention accelerates reasoning by adding a TPV to the model's hidden state at each generation step. This TPV, learned through linear regression on numerous reasoning trajectories, represents a specific direction within the model's hidden space that correlates with advancing through a thought process\cite{Eisenstadt2025Overclocking}. By nudging the hidden state along this vector, the model is guided into a state that it typically associates with a later stage of reasoning, prompting it to bypass intermediate steps and generate a more direct path to the solution.

\subsection{Dynamic Alpha Control Strategies}
\label{sec:dynamic-alpha}

Instead of a constant $\alpha$, we define a token-wise method
\[
\alpha_t = H(\cdot),
\]
which dynamically adjusts at each reasoning step t. Such approach allows model to slow down when uncertainty is high and accelerates when confidence is strong, thereby striking a more effective balance between efficiency and accuracy. The detailed methodology will be illustrated in the following sections.

\subsubsection{Complexity-Guided Alpha Initialization (CG-$\alpha$I)}

This dynamic strategy assigns a global $\alpha$ according to the estimated difficulty of the input. 
To this end, we employ a lightweight \emph{complexity router}, implemented as a small language model (Qwen-4B-instructed\cite{qwen3technicalreport}), 
which is prompted in a few-shot manner to classify each problem $Q$ into three levels: \emph{easy}, \emph{medium}, or \emph{hard}. 
The router can operate either on annotated datasets (MATH 500\cite{lightman2023lets}) or on proxy signals 
such as baseline reasoning length.

Formally, we map the predicted difficulty to an initial $\alpha$ value:
\begin{equation}\label{eq:cg-alpha}
\alpha_{\text{init}} = f_c(\text{difficulty}(Q)) 
\end{equation}
where the mapping assigns larger values to easier problems and smaller values to harder ones, for instance,
\[
\alpha_{\text{init}} = 
\begin{cases}
\alpha_{\text{high}} = 50 & \text{if easy}, \\
\alpha_{\text{mid}} = 30  & \text{if medium}, \\
\alpha_{\text{low}} = 10  & \text{if hard}
\end{cases}
\]

In this way, simpler problems are processed with stronger intervention, while more challenging ones are handled with greater caution.

\subsubsection{Uncertainty-Aware Alpha Scheduling (UA-$\alpha$S)}

Building upon the initialization from CG-$\alpha$I, we further introduce a reactive strategy which adjusts $\alpha$ step by step according to predictive uncertainty. At token $t$, we compute normalized entropy as follows:
\begin{equation}\label{eq:uncertainty}
u_t = \frac{\text{Entropy}(P(\text{token}_t|\cdot))}{\log |V|} 
\end{equation}
where $|V|$ is the vocabulary size.  

We then define the dynamic control function $H(\cdot)$ by mapping $u_t$ to $\alpha_t$ through a sigmoid-shaped transformation:  
\begin{equation}\label{eq:ua-alpha}
\begin{aligned}
\alpha_t &= H(u_t,\alpha_{\text{base}}) \\
&= \alpha_{\text{base}} + (\alpha_{\text{max}} - \alpha_{\text{base}}) \cdot \Big(1 - \sigma\big(k \cdot (u_t - u_{\text{thr}})\big)\Big),
\end{aligned}
\end{equation}
where $\alpha_{\text{base}}$ denotes the baseline intervention strength and $\alpha_{\text{max}}$ is the maximum value.  
In general, both $\alpha_{\text{base}}$ and $\alpha_{\text{max}}$ can be \textbf{flexibly specified depending on the model and application scenario}.  
The hyperparameter $u_{\text{thr}}$ sets the uncertainty threshold, while $k$ controls the sharpness of the transition.  
Accordingly, the hidden state at step $t$ is updated as:
\begin{equation}\label{eq:ua-intervention}
h_t' = h_t + \alpha_t \theta .
\end{equation}
This formulation makes explicit the connection between the abstract update rule $\alpha_t = f(\cdot)$ and its concrete instantiation.  

\subsubsection{Hybrid Adaptive Control (HAC)}  

HAC integrates the two components into a unified mechanism.  
Instead of randomly setting the initial and maximum strengths, HAC first employs the complexity router to determine an initial value $\alpha_{\text{init}}$ (Eq.~\eqref{eq:cg-alpha}), which subsequently serves as the base strength $\alpha_{\text{base}}$ in Eq.~\eqref{eq:ua-alpha}.  
To enable dynamic refinement, we further define the maximum strength as
\[
\alpha_{\text{max}} = \alpha_{\text{init}} + \delta,
\]
where $\delta$ is a user-defined range hyperparameter (typically set to $40$).

The resulting unified control function is expressed as:
\begin{equation}\label{eq:final-alpha}
\alpha_t = H(\alpha_{\text{init}}, u_t) .
\end{equation}

This hybrid design achieves \textit{global alignment} with problem difficulty through $\alpha_{\text{init}}$, while simultaneously enabling \textit{local adaptation} to uncertainty through $u_t$.  
In this way, HAC provides a principled mechanism for dynamic overclocking that balances stability and flexibility.  

\begin{table*}[t]
  \caption{\textbf{Results for DeepSeek-R1-Distill-Llama-8B} on GSM-8K, Math500, and SVAMP at different context lengths (512, 1024, 2048 for Math500; 512, 1024 for GSM-8K; 512, 1024 for SVAMP), showing \#Correct '\#Cr', \#Answered '\#An' and \#Ended '\#En' for each. \textbf{Bold} numbers indicate the best result, and \underline{underline} numbers indicate the second best.}
  \vspace{-4pt}
  \label{tab:model1_results}
  \begin{center}
  \resizebox{0.95\textwidth}{!}{
    \begin{tabular}{@{}l@{~~~}
      ccc@{~~}ccc@{~~~}ccc@{~~~}
      ccc@{~~}ccc@{~~~}ccc@{}}
    \toprule
    \multirow{3}{*}{\textbf{Method}}
      & \multicolumn{6}{c}{\textbf{Math500}}
      & \multicolumn{6}{c}{\textbf{GSM-8K}}
      & \multicolumn{6}{c}{\textbf{SVAMP}} \\
    \cmidrule(lr){2-7} \cmidrule(lr){8-13} \cmidrule(lr){14-19}
      & \multicolumn{3}{c}{\textbf{512}}
      & \multicolumn{3}{c}{\textbf{1024}}
      & \multicolumn{3}{c}{\textbf{512}}
      & \multicolumn{3}{c}{\textbf{1024}}
      & \multicolumn{3}{c}{\textbf{512}}
      & \multicolumn{3}{c}{\textbf{1024}} \\
    \cmidrule(lr){2-4} \cmidrule(lr){5-7}
    \cmidrule(lr){8-10} \cmidrule(lr){11-13}
    \cmidrule(lr){14-16} \cmidrule(lr){17-19}
      & \#Cr & \#An & \#En
      & \#Cr & \#An & \#En
      & \#Cr & \#An & \#En
      & \#Cr & \#An & \#En
      & \#Cr & \#An & \#En
      & \#Cr & \#An & \#En \\
    \midrule
    Llama-8B\cite{guo2025deepseek}          & 0  & 2  & 0  & 60  & 71  & 28
                  & 23  & 23  & 2  & 174  & 183  & 130
                  & \underline{33}& \underline{33} & \underline{11}  & 185& 188& 160\\
    TPV\cite{Eisenstadt2025Overclocking} \(\alpha=50\) & 0  & 1  & 0  & 69  & 77  & 27
                  & \underline{28}  & \underline{28}  & 3  & \underline{192}  & \underline{203}  & \underline{145}
                  & 26  & 26  & 3& \underline{202}& \underline{207}& \underline{178}\\
    TPV\cite{Eisenstadt2025Overclocking} \(\alpha=100\) & 0 & \underline{8} & \underline{3}  & \underline{73}  & \underline{85}  & \underline{29}
                  & 15  & 18  & \textbf{11}  & 86  & 96  & 77
                  & 32& 32& \textbf{22}& 68& 69& 62\\
    Ours          & \textbf{9} & \textbf{9} & \textbf{4}
                  & \textbf{85} & \textbf{98} & \textbf{35}
                  & \textbf{30} & \textbf{30} & \underline{7}
                  & \textbf{199} & \textbf{214} & \textbf{146}
                  & \textbf{38}& \textbf{39}& 8
                  & \textbf{210} & \textbf{214} & \textbf{179} \\
    \bottomrule
    \end{tabular}
  }
  \end{center}
\end{table*}

\begin{table*}[t]
  \caption{\textbf{Results for DeepSeek-R1-Distill-Qwen-32B} on GSM-8K, Math500, and SVAMP at different context lengths (512, 1024 for Math500; 512, 1024 for GSM-8K; 512, 1024 for SVAMP), showing \#Correct '\#Cr', \#Answered '\#An' and \#Ended '\#En' for each. }
  \vspace{-4pt}
  \label{tab:model2_results}
  \begin{center}
  \resizebox{0.95\textwidth}{!}{
    \begin{tabular}{@{}l@{~~~}
      ccc@{~~}ccc@{~~~}ccc@{~~~}
      ccc@{~~}ccc@{~~~}ccc@{}}
    \toprule
    \multirow{3}{*}{\textbf{Method}}
      & \multicolumn{6}{c}{\textbf{Math500}}
      & \multicolumn{6}{c}{\textbf{GSM-8K}}
      & \multicolumn{6}{c}{\textbf{SVAMP}} \\
    \cmidrule(lr){2-7} \cmidrule(lr){8-13} \cmidrule(lr){14-19}
      & \multicolumn{3}{c}{\textbf{512}}
      & \multicolumn{3}{c}{\textbf{1024}}
      & \multicolumn{3}{c}{\textbf{512}}
      & \multicolumn{3}{c}{\textbf{1024}}
      & \multicolumn{3}{c}{\textbf{512}}
      & \multicolumn{3}{c}{\textbf{1024}} \\
    \cmidrule(lr){2-4} \cmidrule(lr){5-7}
    \cmidrule(lr){8-10} \cmidrule(lr){11-13}
    \cmidrule(lr){14-16} \cmidrule(lr){17-19}
      & \#Cr & \#An & \#En
      & \#Cr & \#An & \#En
      & \#Cr & \#An & \#En
      & \#Cr & \#An & \#En
      & \#Cr & \#An & \#En
      & \#Cr & \#An & \#En \\
    \midrule
    Qwen-32B\cite{guo2025deepseek}          & 1& 1& 0& 75& 87& 29
                  & 63  & 73  & 52  & 215  & 212  & 198
                  & 225& 239& 230& 262& 278& 273\\
    TPV\cite{Eisenstadt2025Overclocking} \(\alpha=50\) &5 &\underline{5} &\textbf{2} &89 &105 &\textbf{45}
                  & 79& 87& 68& \underline{234}& 250& 193& 242& 261& 254& \underline{267} & \underline{288}& 282\\
    TPV\cite{Eisenstadt2025Overclocking} \(\alpha=100\) &\underline{8} &\textbf{9} &\underline{1} & \underline{95}& \underline{107}& \underline{43}
                  & \underline{94}  & \underline{108}  & \textbf{78}  & 230  & \underline{253}  & \textbf{209}
                  & \underline{249}& \textbf{277}& \textbf{275} & 264& \textbf{292}& \textbf{289}\\
    Ours          & \textbf{9}& \textbf{9}& \underline{1}& \textbf{102}& \textbf{118}& 36
                  & \textbf{97}& \textbf{112}& \underline{72} & \textbf{245}& \textbf{260}
                  & \underline{208} & \textbf{252} & \underline{272} & \underline{268} & \textbf{272}& \textbf{292} & \underline{286} \\
    \bottomrule
    \end{tabular}
  }
  \end{center}
\end{table*}

\section{Experiments}
\label{sec:print}

In this section, we combine dynamic TPV with two open-sourced models: DeepSeek-R1-Distill-Qwen-32B, DeepSeek-R1-Distill-Llama-8B \cite{guo2025deepseek}and conducted comprehensive token-limited experiments within three widely-used mathematical datasets compared with original model, static TPV to validate our framework's efficiency and effectiveness. All baseline model's are either well finetuned or followed their best parameter settings reported in original papers\cite{Eisenstadt2025Overclocking}.

\subsection{Datasets and Evaluation Metrics}
We evaluate on three mathematical reasoning benchmarks: GSM-8K \cite{cobbe2021gsm8k}, Math500 \cite{lightman2023lets}, and SVAMP \cite{patel2021nlp}. GSM-8K consists of 8.5K grade school word problems focusing on arithmetic reasoning. Math500 is a curated subset of 500 competition-level problems from MATH, spanning algebra, geometry, and other domains. SVAMP is a robustness-oriented dataset derived from elementary math problems through paraphrasing and perturbations. Following prior static TPV settings, we adopt similar training splits for vector initialization, using 80 problems from Math500 with the remaining 420 for evaluation, and 30 problems each from GSM-8K and SVAMP with 300 reserved for evaluation.  

Model performance is assessed using three complementary metrics: \textbf{\#Correct}, the number of correctly solved problems identified via the \texttt{\textbackslash boxed\{\}} expression; \textbf{\#Answered}, the number of generations producing an explicit answer; and \textbf{\#Ended}, the number of generations that finish naturally before hitting the token limit.

\subsection{Dynamic TPV Results}
As shown in Table~2, our dynamic TPV consistently achieves the best \#Cr results under both 512- and 1024-token settings. While fixed-$\alpha$ variants ($\alpha=50,100$) bring certain gains over the base model, their acceleration remains limited and often unstable. In contrast, our adaptive strategy yields clear and consistent improvements: on Math500, dynamic TPV reaches $9/9/4$ under 512 tokens and $85/98/35$ under 1024 tokens, substantially surpassing all baselines. On GSM-8K, it achieves $30/30/7$ Correct under 512 tokens and $199/214/146$ under 1024 tokens, compared with $23$--$94$ Correct for non-adaptive settings. On SVAMP, dynamic TPV further improves to $252/272$ Correct, again outperforming both fixed-$\alpha$ and the base model.

A key advantage of our method is its ability to slow down when necessary. Dynamic TPV flexibly allocates more steps to harder problems, boosting accuracy without sacrificing overall efficiency, as evidenced by its competitive \#Ended and \#Answered counts. This balance between exploration and control prevents both excessive jumps on challenging tasks and stagnation on simpler ones, allowing reasoning to proceed at an appropriate pace. More broadly, dynamic TPV turns static acceleration into a feedback-driven regulation mechanism, aligning the model’s trajectory with task difficulty and confidence signals, and thereby delivering consistently superior performance across datasets and token budgets.

\subsection{Ablation Studies}

 We perform ablation studies on Math500, GSM-8K, and SVAMP under the 1024-token setting (Table~\ref{tab:ablation_single_col}) to evaluate the contributions of CG-$\alpha$I and UA-$\alpha$S. We can observe that removing CG-$\alpha$I substantially reduces Correct scores, highlighting that coarse-grained $\alpha$ intervention is crucial for steering the reasoning trajectory in complex problems. Excluding UA-$\alpha$S causes moderate but consistent drops, suggesting that fine-grained, uncertainty-adaptive $\alpha$ is key to preserving precision in challenging or ambiguous cases. 
 
 The full HAC model, integrating CG-$\alpha$I and UA-$\alpha$S, achieves the highest Correct, Answered, and Ended counts across all datasets, highlighting the complementary nature of coarse- and fine-grained interventions and their collective contribution to effective reasoning.

\begin{table}[t]
  \caption{We perform ablation studies of model components on three datasets (based on DeepSeek-R1-Distill-Qwen-32B) under the 1024 token limit.}
  \vspace{-4pt}
  \label{tab:ablation_single_col}
  \begin{center}
  \resizebox{\columnwidth}{!}{%
    \begin{tabular}{@{}l@{\quad}ccc@{\quad}ccc@{\quad}ccc@{}}
    \toprule
    \multirow{2}{*}{\textbf{Method}} & \multicolumn{3}{c}{\textbf{Math500}} & \multicolumn{3}{c}{\textbf{GSM-8K}} & \multicolumn{3}{c}{\textbf{SVAMP}} \\
    \cmidrule(lr){2-4} \cmidrule(lr){5-7} \cmidrule(lr){8-10}
    & \#Cr & \#An & \#En & \#Cr & \#An & \#En & \#Cr & \#An & \#En \\
    \midrule
    Base Model & 75 & 87 & 29 & 215 & 212 & 198 & 262 & 278 & 273 \\
    \midrule
    HAC (w/o CG-$\alpha$I) & 94& 109& 38& 238& 253& 211& 266 & 281& 276\\
    HAC (w/o UA-$\alpha$S) & 79& 94& 34& 229& 246& 191& 269& 288& 285\\
    \textbf{HAC (Ours)} & \textbf{102} & \textbf{118} & \textbf{36} & \textbf{245} & \textbf{260} & \textbf{208} & \textbf{272}&\textbf{292}& \textbf{286}\\
    \bottomrule
    \end{tabular}%
  }
  \end{center}
\end{table}

\section{Conclusion}
\label{sec:conclusion}

In this work, we introduce Adaptive Overclocking, a dynamic, closed-loop control system that elevates static intervention to enhance LLM reasoning efficiency. Our method modulates reasoning speed using two complementary signals: a static initialization based on problem complexity (CG-$\alpha$I), followed by real-time adjustments guided by token-level predictive uncertainty. This state-aware mechanism operates entirely at inference time, requires no model retraining, and can be seamlessly combined with existing methods. Extensive experiments demonstrate that our adaptive approach consistently outperforms static baselines, achieving a superior trade-off between accuracy and computational efficiency. This work validates the efficacy of controlling LLMs via their internal states and suggests future research avenues, such as learned control policies and richer signal integration, paving the way for more powerful, efficient, and adaptable models.
\vfill\pagebreak

\bibliographystyle{IEEEbib}
\bibliography{strings,refs}

\end{document}